\documentclass{article}
\usepackage{spconf,amsmath,epsfig}
\usepackage{graphicx}
\usepackage{diagbox}
\usepackage{stackengine}
\usepackage{multirow}
\usepackage{booktabs}
\usepackage{verbatim}
\usepackage{lipsum,multicol,times, epsfig, graphicx, amsmath, amssymb,  booktabs, url, multirow, comment, subfigure, mwe}
\usepackage[colorlinks]{hyperref}
\usepackage[normalem]{ulem}
\graphicspath{ {./images/} }

\title{Detecting Urban Changes with Recurrent Neural Networks from Multitemporal Sentinel-2 Data}
\name{Maria Papadomanolaki\textsuperscript{1,2*}, Sagar Verma\textsuperscript{2,3*}, Maria Vakalopoulou\textsuperscript{2}, Siddharth Gupta\textsuperscript{3}, Konstantinos Karantzalos\textsuperscript{1}}
\address{%
$^{1}$ \quad Remote Sensing Laboratory, National Technical University of Athens, Greece\\
$^{2}$ \quad CVN, CentraleSup\'elec, Universit\'e Paris-Saclay and INRIA Saclay, France\\
$^{3}$ \quad Granular AI, MA, USA \\
mar.papadomanolaki@gmail.com, \{sagar.verma,maria.vakalopoulou\}@centralesupelec.fr,  \\
sid@granular.ai, karank@central.ntua.gr
}

\begin{document}
%
\maketitle
\begin{abstract}
The advent of multitemporal high resolution data, like the Copernicus Sentinel-2, has enhanced significantly the potential of monitoring the earth's surface and environmental dynamics. In this paper, we present a novel deep learning framework for urban change detection which combines state-of-the-art fully convolutional networks (similar to U-Net) for feature representation and powerful recurrent networks (such as LSTMs) for temporal modeling. We report our results on the recently publicly available bi-temporal Onera Satellite Change Detection (OSCD) Sentinel-2 dataset, enhancing the temporal information with additional images of the same region on different dates. Moreover, we evaluate the performance of the recurrent networks as well as the use of the additional dates on the unseen test-set using an ensemble cross-validation strategy. All the developed models during the validation phase have scored an overall accuracy of more than 95\%, while the use of LSTMs and further temporal information, boost the F1 rate of the change class by an additional 1.5\%.
\end{abstract}
\section{Introduction}
\label{sec:intro}

\begin{figure}[t]
    \centering
        \stackunder[6pt]{\includegraphics[scale=0.35]{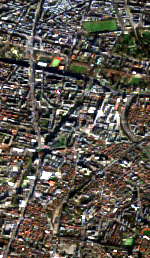}}{}
         \stackunder[6pt]{\includegraphics[scale=0.35]{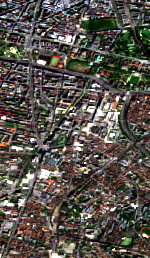}}{}
        \stackunder[6pt]{\includegraphics[scale=0.35]{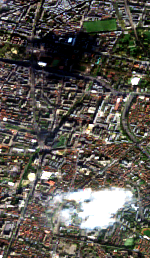}}{}
        \stackunder[6pt]{\includegraphics[scale=0.35]{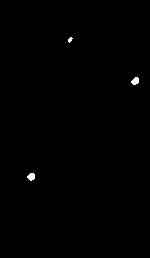}}{} \\    \vspace{-2mm}
        \stackunder[6pt]{\includegraphics[scale=0.35]{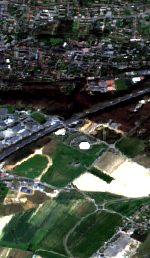}}{Date 1}
        \stackunder[6pt]{\includegraphics[scale=0.35]{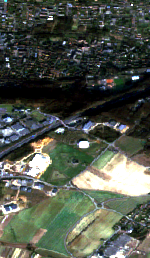}}{Date 3}
        \stackunder[6pt]{\includegraphics[scale=0.35]{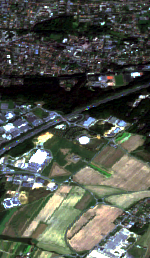}}{Date 5}
        \stackunder[6pt]{\includegraphics[scale=0.35]{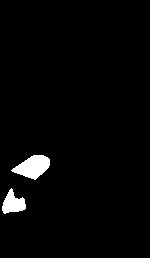}}{Change Mask}        \\
        \vspace{-3mm}
     \caption{Learning specific types of change such as urbanisation is a challenging task. The concurrent exploitation of more than two temporal images is more robust in detecting accurately changes since the presence of clouds/shadows (top) or vegetation/agricultural alterations may be significant (bottom).}
     \label{fig:teaser}
\end{figure}

Change detection is a critical issue for the remote sensing community as it provides an effective way of monitoring the globe. A thorough understanding of the earth's land usage and land cover (LULC) in time can be made by modeling the changes that occur owing to man-made structures and natural phenomena. As far as human intervention on earth is concerned, change detection techniques offer valuable information on a variety of topics such as urban sprawl, water and air contamination levels, illegal constructions, etc. In this way, we can fully comprehend the future LULC tendencies, take precautions and design more appropriate city infrastructures. \let\thefootnote\relax\footnotetext{* Authors with equal contribution}

However, even if nowadays we can have access to a large amount of multitemporal datasets provided by satellites such as Landsat and Sentinel, the problem of change detection is very challenging. Traditional methods, summarised in surveys such as~\cite{rak:05}, use handcrafted techniques which heavily rely on pre-processing and post-processing making them not so easily adaptive to images that cover large areas. Change detection is a non-trivial problem as the accuracy of a method is highly influenced by registration errors~\cite{vkk:15} and illumination changes that do not really correspond to semantic changes.

Recently, with the advances in deep learning-based methods in different fields, a variety of change detection techniques have been proposed. In particular, in~\cite{mou2018sst} a deep patch-based architecture is proposed where bi-temporal patches are processed in parallel by a series of dilated convolutional layers generating features which are then fed to a recurrent sub-network to learn sequential information. In the end, fully-connected layers are used to create the change prediction map. Although patch-based techniques produce promising results, they are time-consuming since they need to process every single pixel of the image individually. Recently, Daudt et al.~\cite{daudt2018icip} suggested three different fully-convolutional siamese networks based on the U-Net architecture~\cite{olaf2015miccai} aiming to address this problem and detect accurately the regions with changes. However, this approach is lacking the appropriate modeling of the data's temporal pattern.

In this paper, we investigate the use of recurrent networks and in particular fully convolutional Long Short-Term Memory (LSTM)~\cite{sepp1997neural} layers for pixel-wise detection of urbanization from multi-temporal high-resolution data. The proposed deep learning model uses a simple U-Net architecture to compute spatial features from multi-date inputs while LSTM blocks learn the temporal change pattern. Our experiments were based on the recently publicly available Onera Satellite Change Detection dataset (OSCD)~\cite{daudt2018igarss} after enriching it with more temporal information for each of the provided cities. Our final aim is to study the behaviour and prospects of such a model when multitemporal data are available eliminating at the same time the need for any fully-connected layers.

\vspace{-0.3cm}
\section{Methodology}
\begin{figure}
\begin{center}
\includegraphics[width=8cm,height=6cm]{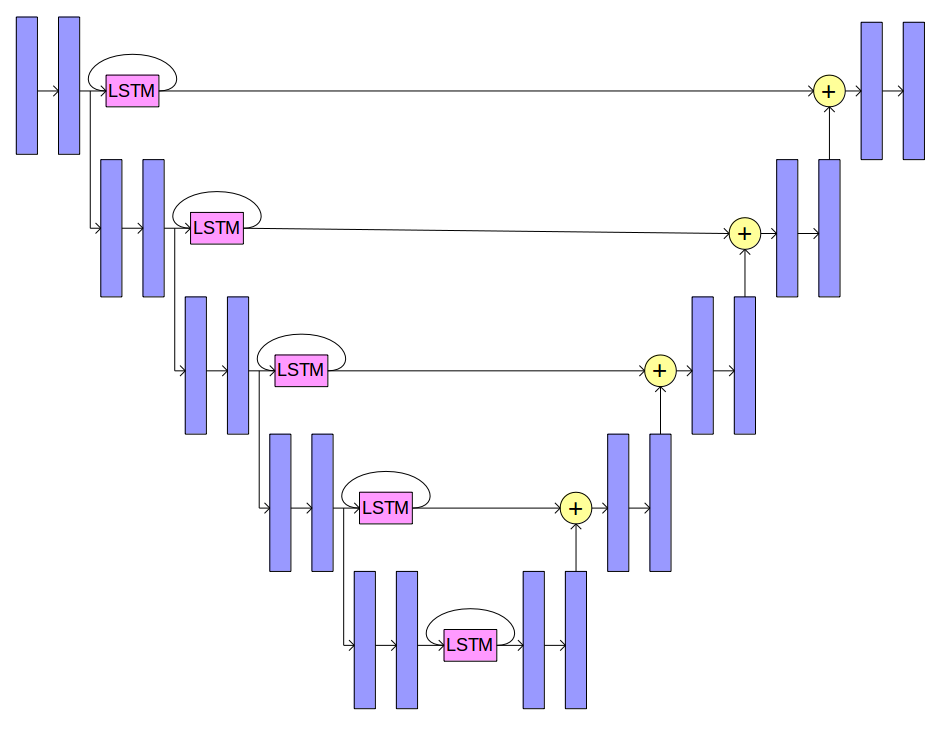}
\end{center}
\vspace{-0.8cm}
\caption{Graphical illustration of the employed architecture. The left and the right part demonstrate the network's encoder and decoder respectively. Blue layers represent convolutional blocks while pink boxes illustrate the recurrent processing layers.}
\label{fig:network}
\end{figure}

The employed network is based on the U-Net architecture and it is demonstrated in Figure~\ref{fig:network}. Each block involves convolutional, batch normalization and rectified linear unit (ReLU) activations. All  the convolutional operations apply 3x3 filters with both stride and padding being equal to 1. As far as the encoder is concerned, the first convolutional block increases the depth to 16, while the height and width of the input volume remain unchanged. The following four blocks follow the same pattern increasing the depth to twice its size and including also a 2x2 max pooling operation. As a result, at the end of the encoder the input has been downsampled at one fourth of its original dimension containing 256 planes.

The encoding process is repeated for every different date independently while at the same time, recurrent blocks existing in each of the five encoding levels calculate the temporal relationship among the outputs. This is implemented by replacing the standard fully-connected LSTM operations with convolutional structures. In practice, the recurrent operations' weights are no longer simple matrices, but convolutional layers which constitute in this way an end-to-end trainable framework.

Next, the decoder receives the encoder's last temporal volume to upsample it back to its original dimensions. For this purpose, five convolutional blocks similar to the encoder are used, this time applying 2x2 upsampling operations instead of max-pooling ones. In addition, the resulted feature map of each upsampling operation is concatenated with the calculated temporal pattern of the symmetrical block existing in the encoder part. In this way, higher resolution information is combined with lower resolution information producing more sophisticated features and maintaining spatial and temporal knowledge. Finally, at the end of the model, a 1x1 convolution operation is applied to compute the final probability heatmap detecting areas of urbanisation.

\subsection{Dataset and Implementation Details}
All the experiments were conducted using the Onera Satellite Change Detection dataset (OSCD)~\cite{daudt2018igarss} which consists of Sentinel-2 satellite images depicting 24 different cities around the world for two distinct dates. 13 spectral channels are available for each image pair with ground truth information provided for 14 cities. Our setup follows the submission system guidelines\footnote{http://dase.grss-ieee.org} where the 14 image pairs are used for training and the rest for testing.

\begin{table*}[]
    \centering
    \begin{tabular}{c c c c c c c c c}
        \toprule
         Abudhabi & Beirut & Chongqing & Dubai & Hong Kong & Milano & Paris & Rio\\
         \midrule
         2016/01/20 & 2015/08/20 &  2017/04/14 & 2015/12/11 & 2016/09/27 & 2016/12/28 & 2016/11/30 & 2016/04/24\\
         2016/09/29 & 2015/12/08 &  2017/07/23 & 2016/06/08 & 2017/01/25 & 2017/05/27 & 2017/02/15 & 2017/02/18\\
         2017/03/18 & 2016/04/26 &  2017/09/16 & 2016/11/05 & 2017/04/02 & 2017/08/15 & 2017/04/09 & 2017/05/09\\
         2017/09/09 & 2017/04/21 &  2018/01/14 & 2017/06/03 & 2017/10/22 & 2017/11/18 & 2017/08/29 & 2017/07/28\\
         2018/03/28 & 2017/10/03 &  2018/04/02 & 2018/03/30 & 2018/03/23 & 2018/01/22 & 2017/11/07 & 2017/10/11\\
         \bottomrule
    \end{tabular}
    \vspace{-0.1cm}
    \caption{Acquired dates for some cities to enhance the already provided OSCD bi-temporal information.}
    \label{tab:multidate}
\end{table*}

As mentioned earlier, additional Sentinel-2 images depicting the provided cities in different dates were acquired to further enrich the temporal information of the OSCD dataset. In Table~\ref{tab:multidate} we present some of the additional selected images for 8 different cities.  The first and last rows include the \textit{before} and \textit{after} dates that are already provided in the OSCD dataset. In general, we tried to obtain dates corresponding to similar seasons for every city, adapting them as much as possible to the two existing OSCD dates.

For the training process, patches of size 32x32 were produced with a stride of either 6 in case \textit{change} pixels were included, or 32 in case the patch did not include pixels for the \textit{change} class. This strategy was applied as a data augmentation approach to enrich the training samples that contain \textit{change}. In addition, more data augmentation techniques mainly used from the computer vision community, namely flipping in all possible angles proportional to 90 degrees, were implemented for patches whose number of \textit{change} pixels exceeded the threshold of 5\% for the entire patch. Lastly, each class was associated with a weight inversely proportional to the total pixel number included in it. A total of 32421 patches containing both \textit{change} and \textit{non-change} pixels resulted from the 14 training cities and was feedforwarded to the proposed architecture for training. To provide more robust results, our final predictions were produced by an ensemble of different trained models following a cross-validation. Giving some more details, the training patches were divided into five equal parts and the same model was trained five times using all possible combinations of the dataset partitions. Then, predictions for the testing images were produced from all five models and in the end the final predictions were formulated by averaging the five model outcomes.
Regarding hyperparameter details, the chosen optimizer was Adam while batch size and learning rate values were equal to 64 and 0.0001 respectively. All experimental setups were conducted using the PyTorch deep learning library~\cite{paszke2017nips} on a single NVIDIA GeForce GTX TITAN with 12 GB of GPU memory, with the training time for each model being approximately 70 minutes and the testing for an entire city approximately 1 minute.

\vspace{-0.3cm}
\section{Experimental Results and Discussion}
\begin{table}[t!]
\scalebox{0.75}{
\centering
 \begin{tabular}{c c c c c c}
 \hline
 \toprule
 Architecture & \#Dates/ \#Channels & Precision & Recall & OA & F1 \\
\midrule
 \multirow{4}{*}{U-Net} & 2/13 & 60.80 & 50.51 & 95.76 & 55.18 \\
 & 2/4  & 58.87 & 53.90 & 95.67 & 56.28 \\
 & 3/4  & 59.28 & 51.94 & 95.67 & 55.36 \\
 & 5/4  & 61.21 & 47.57 & 95.73 & 53.53 \\
 \midrule
 \multirow{3}{*}{U-Net + LSTM} & 2/4  & 59.88 & \textbf{54.78} & 95.77 & 57.22 \\
 & 3/4  & 60.66 & 50.91 & 95.76 & 55.36 \\
 & 5/4  & \textbf{63.59} & 52.93 & \textbf{96.00} & \textbf{57.78} \\ [1ex]
 \bottomrule
 \end{tabular}}
 \vspace{-2mm}
 \caption{Quantitative evaluation of the proposed framework. Precision, recall and F1 rates are associated to the \textit{change} class. All the rows except the first one demonstate results using the RGB-NIR bands.}
\label{table:accuracies}
\end{table}

\begin{figure*}[h]
    \centering
  \begin{minipage}[t]{.16\linewidth}
    \centering
    \includegraphics[width=2.8cm, height=2.8cm]{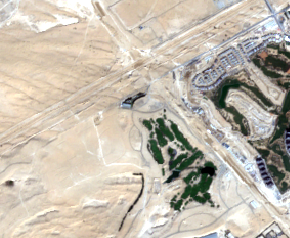}
    \label{f1:dubai1}
  \end{minipage}
  \hfill
  \begin{minipage}[t]{.16\linewidth}
    \centering
    \includegraphics[width=2.8cm, height=2.8cm]{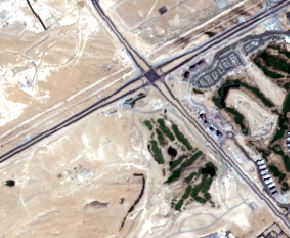}
    \label{f1:dubai2}
  \end{minipage}
  \hfill
  \begin{minipage}[t]{.16\linewidth}
    \centering
    \includegraphics[width=2.8cm, height=2.8cm]{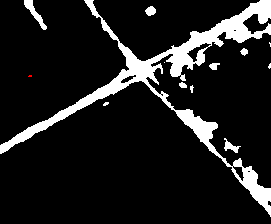}
    \label{f1:dubai2}
  \end{minipage}
  \hfill
  \begin{minipage}[t]{.16\linewidth}
    \centering
    \includegraphics[width=2.8cm, height=2.8cm]{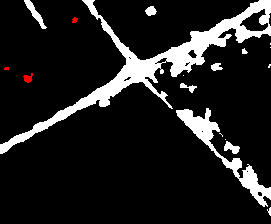}
    \label{f1:dubai2}
  \end{minipage}
  \hfill
  \begin{minipage}[t]{.16\linewidth}
    \centering
    \includegraphics[width=2.8cm, height=2.8cm]{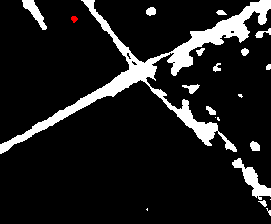}
    \label{f1:dubai2}
  \end{minipage}
  \hfill
  \begin{minipage}[t]{.16\linewidth}
    \centering
    \includegraphics[width=2.8cm, height=2.8cm]{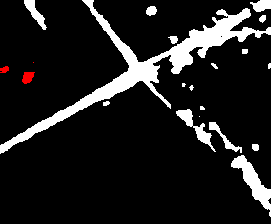}
    \label{f1:dubai2}
  \end{minipage}
 \\ \vspace*{-0.3cm}
  \begin{minipage}[t]{.16\linewidth}
    \centering
    \includegraphics[width=2.8cm, height=2.8cm]{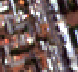}
    \label{f1:dubai1}
  \end{minipage}
  \hfill
  \begin{minipage}[t]{.16\linewidth}
    \centering
    \includegraphics[width=2.8cm, height=2.8cm]{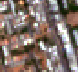}
    \label{f1:dubai2}
  \end{minipage}
  \hfill
  \begin{minipage}[t]{.16\linewidth}
    \centering
    \includegraphics[width=2.8cm, height=2.8cm]{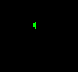}
    \label{f1:dubai2}
  \end{minipage}
  \hfill
  \begin{minipage}[t]{.16\linewidth}
    \centering
    \includegraphics[width=2.8cm, height=2.8cm]{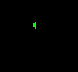}
    \label{f1:dubai2}
  \end{minipage}
  \hfill
  \begin{minipage}[t]{.16\linewidth}
    \centering
    \includegraphics[width=2.8cm, height=2.8cm]{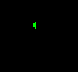}
    \label{f1:dubai2}
  \end{minipage}
  \hfill
  \begin{minipage}[t]{.16\linewidth}
    \centering
    \includegraphics[width=2.8cm, height=2.8cm]{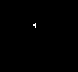}
    \label{f1:dubai2}
  \end{minipage}
 \\ \vspace*{-0.3cm}
  \begin{minipage}[t]{.16\linewidth}
    \centering
    \includegraphics[width=2.8cm, height=2.8cm]{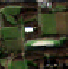}
    \label{f1:dubai1}
  \end{minipage}
  \hfill
  \begin{minipage}[t]{.16\linewidth}
    \centering
    \includegraphics[width=2.8cm, height=2.8cm]{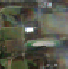}
    \label{f1:dubai2}
  \end{minipage}
  \hfill
  \begin{minipage}[t]{.16\linewidth}
    \centering
    \includegraphics[width=2.8cm, height=2.8cm]{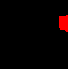}
    \label{f1:dubai2}
  \end{minipage}
  \hfill
  \begin{minipage}[t]{.16\linewidth}
    \centering
    \includegraphics[width=2.8cm, height=2.8cm]{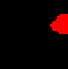}
    \label{f1:dubai2}
  \end{minipage}
  \hfill
  \begin{minipage}[t]{.16\linewidth}
    \centering
    \includegraphics[width=2.8cm, height=2.8cm]{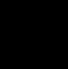}
    \label{f1:dubai2}
  \end{minipage}
  \hfill
  \begin{minipage}[t]{.16\linewidth}
    \centering
    \includegraphics[width=2.8cm, height=2.8cm]{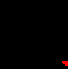}
    \label{f1:dubai2}
  \end{minipage}
 \\ \vspace*{-0.3cm}
  \begin{minipage}[t]{.16\linewidth}
    \centering
    \includegraphics[width=2.8cm, height=2.8cm]{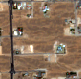}
    \label{f1:dubai1}
  \end{minipage}
  \hfill
  \begin{minipage}[t]{.16\linewidth}
    \centering
    \includegraphics[width=2.8cm, height=2.8cm]{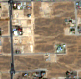}
    \label{f1:dubai2}
  \end{minipage}
  \hfill
  \begin{minipage}[t]{.16\linewidth}
    \centering
    \includegraphics[width=2.8cm, height=2.8cm]{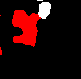}
    \label{f1:dubai2}
  \end{minipage}
  \hfill
  \begin{minipage}[t]{.16\linewidth}
    \centering
    \includegraphics[width=2.8cm, height=2.8cm]{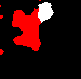}
    \label{f1:dubai2}
  \end{minipage}
  \hfill
  \begin{minipage}[t]{.16\linewidth}
    \centering
    \includegraphics[width=2.8cm, height=2.8cm]{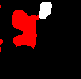}
    \label{f1:dubai2}
  \end{minipage}
  \hfill
  \begin{minipage}[t]{.16\linewidth}
    \centering
    \includegraphics[width=2.8cm, height=2.8cm]{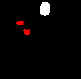}
    \label{f1:dubai2}
  \end{minipage}
\vspace{-0.5cm}
\caption{Qualitative evaluation of the employed frameworks on testing images. From top to bottom: Dubai, Brasilia, Milano, Las Vegas. From left to right: Image1, Image2, 2/4 U-Net, 2/4 U-Net+LSTM, ,3/4 U-Net+LSTM, 5/4 U-Net + LSTM. White: True Positives, Black: True Negatives, Red: False Positives, Green: False Negtatives.}\label{fig:animals}
\label{fig:fig1}
\vspace{-0.5cm}
\end{figure*}

Different model combinations were evaluated on the testing images, the results of which are shown in Table~\ref{table:accuracies}. Precision, recall, F1 score and Overall Accuracy rates have resulted from the confusion matrix of the testing images for each different approach while all quantitative numbers except OA are associated to the \textit{change} class. Beginning with the simple U-Net architecture, judging from the F1 rate we can observe that between 2/13 and 2/4, \textit{change} has been more successfully detected in the latter case. This can be explained by the fact that RGB-NIR channels provide the highest spatial resolution in Sentinel-2 images. Hence, the integration of the remaining lower resolution channels to the training process requires the use of more sophisticated pansharpening methods than only simple resizing. Based on this fact and since 13 channel volumes require much more computational power and registration preprocessing, we based our experiments on the four higher resolution channels.

Continuing with the quantitative analysis, one can notice that for the simple U-Net architecture the addition of temporal dates reduce the performance of the model, indicating that this architecture is not capable to model properly the temporal information that exists in more than two pairs of images. On the other hand, the use of the LSTM blocks prove their significance by reporting their best performance after using all the available temporal information. In addition, recall rates remain over 50\% in all Dates-Channels combinations.

In Figure~\ref{fig:fig1}, we can also present some qualitative results from the different models. Red and green parts in the predictions' visualization represent false positive and false negative areas that we have manually annotated, only for the qualitative analysis of our method, since no ground truth is available for the testing part of the dataset. As we can observe, the LSTM frameworks are able to distinguish clouds from change as shown in the Milano case in the third row. In addition, bi-temporal approaches tend to confuse urban changes with changes related to bare soil lands, whereas multi-date approaches seem to overcome this problem. This is obvious in the last row of Figure~\ref{fig:fig1} where false positives are reduced when more dates are exploited. In the Dubai case however, bare-soil-related false positives continue to exist in the 5-dates case, whereas they are eliminated in the 3-dates case. Such false positive predictions may be a result of various factors like clouds, improper registration etc.

\section{Conclusion}
In this paper, we investigated the behaviour of LSTM convolutional blocks integrated into fully-convolutional deep architectures for urban change detection. Several experiments were implemented using various combinations of architectures and inputs. Results on the OSCD dataset indicate that the use of recurrent networks can boost both F1 and Overall Accuracy rates. Our future steps include the investigation of other types and combinations of recurrent fully convolutional architectures on high-resolution images. Moreover, we will also investigate the use of these models for change detection on very high-resolution satellite imagery.
\bibliographystyle{IEEEbib}
\small{
\bibliography{ms}
}
\end{document}